\documentclass[12pt]{article}

\usepackage[english]{babel}
\usepackage[utf8]{inputenc}
\usepackage[T1]{fontenc}
\usepackage{blindtext}
\usepackage{enumitem}
\usepackage{float}
\usepackage{commath}
\usepackage{etoolbox}
\patchcmd{\subsection}{\bfseries}{\itshape}{}{}
\usepackage{makecell}
\setlist[description]{font=\normalfont\itshape\space\underline}
\usepackage[a4paper,top=3cm,bottom=2cm,left=2cm,right=2cm,marginparwidth=1.75cm]{geometry}

\usepackage{amsmath}
\usepackage{graphicx}
\usepackage[export]{adjustbox}
\usepackage[colorinlistoftodos]{todonotes}
\usepackage{hyperref}
\hypersetup{colorlinks=true,allcolors=black}
\usepackage{natbib}
\usepackage{authblk}
\title{GPU optimization of the 3D Scale-invariant Feature Transform Algorithm and a Novel BRIEF-inspired 3D Fast Descriptor}
\author[a, b]{Jean-Baptiste Carluer}
\author[a]{Laurent Chauvin}
\author[c]{Jie Luo}
\author[c]{William M. Wells III}
\author[c, d]{In\^es Machado}
\author[e]{Rola Harmouche}
\author[a]{Matthew Toews}

\affil[a]{École de technologie supérieure ÉTS, Montreal, Canada}
\affil[b]{Université de Nantes, Nantes, France}
\affil[c]{Brigham and Women's Hospital, Harvard Medical School, Boston, U.S.A.}
\affil[d]{Instituto Superior Técnico, Universidade de Lisboa, Lisbon, Portugal}
\affil[e]{National Research Council, Montreal, Canada}

\date{\today}

\begin{document}
\maketitle
\vspace{0.5cm}
\hrule
\begin{abstract}
This work details a highly efficient implementation of the 3D scale-invariant feature transform (SIFT) algorithm, for the purpose of machine learning from large sets of volumetric medical image data. The primary operations of the 3D SIFT code are implemented on a graphics processing unit (GPU), including convolution, sub-sampling, and 4D peak detection from scale-space pyramids. The performance improvements are quantified in keypoint detection and image-to-image matching experiments, using 3D MRI human brain volumes of different people. Computationally efficient 3D keypoint descriptors are proposed based on the Binary Robust Independent Elementary Feature (BRIEF) code, including a novel descriptor we call Ranked Robust Independent Elementary Features (RRIEF), and compared to the original 3D SIFT-Rank method\citep{toews2013efficient}. The GPU implementation affords a speedup of approximately 7X beyond an optimised CPU implementation, where computation time is reduced from 1.4 seconds to 0.2 seconds for 3D volumes of size (145, 174, 145) voxels with approximately 3000 keypoints. Notable speedups include the convolution operation (20X), 4D peak detection (3X), sub-sampling (3X), and difference-of-Gaussian pyramid construction (2X). Efficient descriptors offer a speedup of 2X and a memory savings of 6X compared to standard SIFT-Rank descriptors, at a cost of reduced numbers of keypoint correspondences, revealing a trade-off between computational efficiency and algorithmic performance. The speedups gained by our implementation will allow for a more efficient analysis on larger data sets. Our optimized GPU implementation of the 3D SIFT-Rank extractor is available at (\url{https://github.com/CarluerJB/3D_SIFT_CUDA}).\\
\end{abstract}
\hrule
\vspace{0.5cm}
\section{Introduction}
The rapid growth of image acquisition and storage technologies have resulted in a wealth of data. Algorithms based on raw image data such as deep convolutional neural networks (CNNs)~\citep{lecun1989backpropagation} leverage such large data sets to obtain high accuracy for tasks such as classification and prediction, and have subsequently been extensively used for such tasks in medical contexts. A central challenge remains efficiently processing large amounts of 3D medical image data, particularly in the case of computational algorithms such as deep neural networks which require large amounts of training data, memory and computational power and are limited by algorithmic complexity, bandwidth and RAM capacity. This motivates the use of sparse, data-efficient image representations such as SIFT keypoints~\citep{Lowe}, which enable machine learning algorithms to manipulate sets of training data that are several orders of magnitude larger than if the raw image data were to be used. This is especially true for analyzing volumetric medical image data, where at the finest grain of analysis each individual subject may represent a unique class. For example, the ability to identify and characterize individuals and family members in large sets 3D scans is crucial in achieving accurate, personalized healthcare and minimizing patient-level labeling errors. Currently the 3D SIFT-Rank keypoint indexing~\citep{toews2013efficient} is the only method reporting this capability~\citep{chauvin2019neuroimage,chauvin2021efficient}.

\begin{figure}[H]
        
        \begin{center}
        \begin{tabular}{cc}
        \includegraphics[width=.4\textwidth]{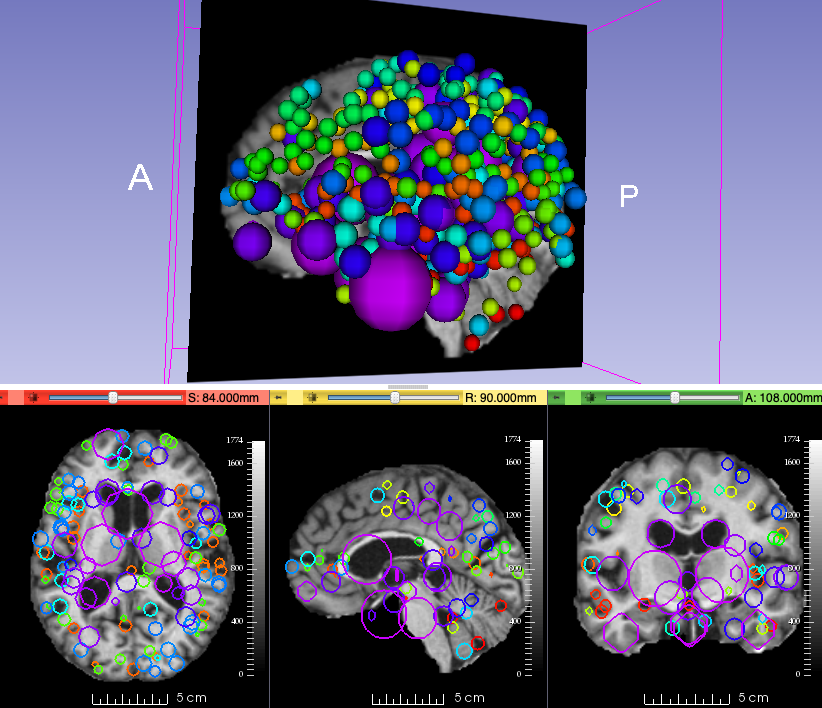} &
        \includegraphics[width=.5\textwidth]{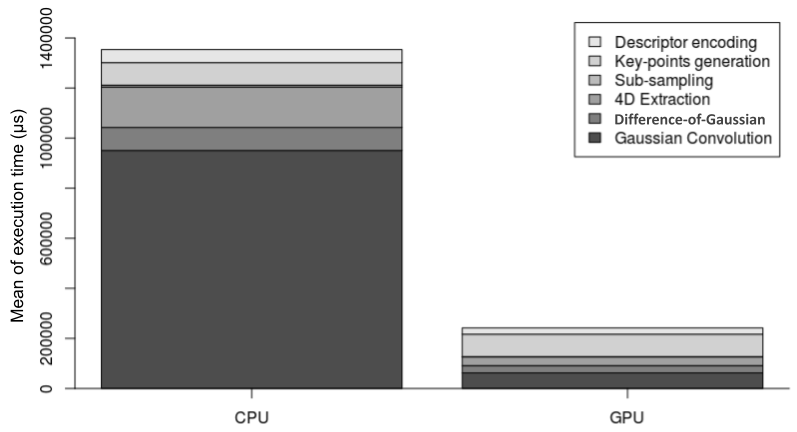}  \\
        (a) 3D Keypoint Example & (b) GPU Speedup \\
        \end{tabular}
        \end{center}
    \caption{(a) Visualization of 3D SIFT keypoints extracted in a human brain MRI from the OASIS dataset~\citep{marcus2007open}, spheres represent the 3D location $\bar{x} \in R^3$ and size or scale $\sigma \in R^1$ of keypoints, color indicates the scale. Note the high degree of left-right symmetry. (b) Total execution time comparison between CPU and GPU implementations. Note a speedup of $7\times$ between CPU (left, previous 3D SIFT-Rank method~\citep{toews2013efficient}) and GPU (right, this work) implementations.}
    \label{fig:Conclusion_times_total}
\end{figure}

In this paper, we describe a novel implementation of the 3D SIFT algorithm using a GPU architecture that shows significant speed improvements. Figure~\ref{fig:Conclusion_times_total} illustrates the result of this work including (a) an example of 3D SIFT keypoints extracted from a volumetric human brain MRI and (b) the $7 \times$ speedup afforded by our GPU implementation. The rest of this paper describes the related work context, then our GPU implementation of the 3D SIFT algorithm and efficient descriptor comparisons. 

\section{Related Work}


This related work section describes the context of keypoint-based 3D medical image analysis, including convolutional neural networks and local invariant keypoint analysis. Readers interested solely in our 3D GPU-SIFT model may continue directly to the following section without loss of information. 

Image-based machine learning can be generally described in the framework of artificial neural networks, where technological advancements typically involve reducing the complexity of neural network training the large parameter space and increasing the ability to generalise from smaller training data sets. Artificial neural networks were developed to model and simulate biological neural network processing, i.e. the perceptron model of Rosenblatt where the output of a neuron is a weighted linear combination of the inputs~\citep{rosenblatt1958perceptron}. 'Deep' networks emerged shortly thereafter with the multi-layer perceptron (MLP)~\citep{rosenblatt1962principles}, where perceptrons were organized into multiple, feed-forward layers evaluated in a sequential fashion. A persistent theme has been the incorporation of mathematical invariance, allowing networks to train from fewer data points, or equivalently to generalize to unseen data. LeCun developed the translation-invariant convolutional neural network (CNN) architecture in the 1980s~\citep{lecun1989backpropagation}, which greatly reduced the memory requirements of the weights from $O(N^2)$ in the number $N$ of pixels for fully-connected MLP network layers to $O(N)$ for sets of weights computed locally and shared across the image in a manner invariant to image translation, i.e. convolution filters. While highly effective, CNN training via the backpropagation algorithm remained too computationally complex for practical applications from sequential CPU-based computers until the development of highly parallel GPU-based convolution allowed training of deep CNNs from 100000s of images~\citep{krizhevsky2012imagenet}, e.g. the ImageNet dataset consisting of 1000 objects with 1000 training examples per category).

Keypoint algorithms were developed to focus on small sets of informative image points rather than entire images. Keypoints are detected via convolutional filters with specific mathematical forms, guaranteeing invariance to input image translation~\citep{harris1988combined},  scaling~\citep{lindeberg1998feature}, similarity~\citep{Lowe} and affine deformations~\citep{mikolajczyk2004scale}. Keypoint methods remained the defacto standard for image-based learning until the development of programmable graphics processing units (GPUs), e.g. the Compute Unified Device Architecture (CUDA) development system, that could be used to train full image-based CNNs via massive parallelization of the backpropagation learning algorithm~\citep{krizhevsky2012imagenet}.  This article focuses on the SIFT algorithm designed for keypoint extraction by David Lowe in 1999~\citep{Lowe}. The backbone of the SIFT algorithm is the Gaussian scale-space~\citep{lindeberg1994scale} generated via recursive Gaussian convolution filtering, it can be generated efficiently via separable and even uniform 1D filters~\citep{wells1986efficient}. Advantages of the SIFT algorithm include the fact that keypoints are identified in a manner invariant to image resolution, object pose, or intensity variations, no training procedure or data are required, Gaussian and Gaussian derivative filters are rotationally symmetric and uniformly sampled and unbiased by the specific training data used. For these reasons, the SIFT keypoint algorithm and descriptor remains and efficient, competitive solution in contexts with few training examples, i.e. matching images of the same specific 3D objects~\citep{bellavia2020there} or scenes~\citep{mishchuk2017working}. The functioning of the SIFT algorithm is analogous to aspects of visual processing within the mammalian vision system, including retinal center-surround processing in the lateral geniculate nucleus and oriented gradient operators organised into hypercolumns~\citep{hubel1968receptive}. The scale-space derived from Gaussian convolution $I(x,y,z,\sigma)=I(x,y,z)*N(\sigma)=$ representing the process of isotropic diffusion of particles (ex. Brownian motion ) or heat. The Laplacian-of-Gaussian $\nabla^2 I(x,y,z,\sigma)$ is a scale-normalized operator analogous to the Laplacian operator forming the kinetic energy component of the Hamiltonian, e.g. as found in the Schrodinger equation.


The SIFT algorithm has been generalized to 3D volumetric image data in a variety of applications including video processing~\citep{scovanner20073}, 3D object detection~\citep{flitton2010object}, and medical image analysis~\citep{cheung2009n,allaire2008full,toews2013efficient,rister2017volumetric}. Our work here is based on the 3D SIFT-Rank approach~\citep{toews2013efficient}, which is unique in the way local keypoint orientation and descriptors are computed. Keypoint orientations may be parameterized via a suitable representation such as quaternions or 3x3 rotation matrices, are generally determined from dominant gradients of the scale-space $\nabla I(x,y,z,\sigma)$ in a scale-normalized neighborhood about the keypoint. The 3D SIFT-Rank method identifies dominant peaks in a discrete spherical histogram of gradient, rather than solid angle histograms~\citep{scovanner20073,allaire2008full}, partial 3D orientation information~\citep{cheung2009n} or principal components of the local gradient\citep{rister2017volumetric}. This allows identification multiple dominant orientations (3x3 rotation matrices) at each keypoint $(x,y,z,\sigma)$, leading to multiple descriptors per keypoint and providing robustness to noise, rather than single orientations per keypoint\citep{rister2017volumetric}. Descriptors generally encode the scale-space gradient $\nabla I(x,y,z,\sigma)$ in neighborhood about the keypoint, following reorientation. The 3D SIFT-Rank method adopts a compact 64-element gradient orientation descriptor, where the local orientation space is sampled according to $x\times y\times z=2\times2\times2=8$ spatial bins and 8 orientation bins. Other approaches adopt descriptors that are orders of magnitude larger, e.g. 768-elements from $4^3$ spatial bins x 12 orientation bins, leading to similar image matching performance at a much greater memory footprint \citep{rister2017volumetric}. Finally, the SIFT-Rank descriptor is normalized by ranking~\citep{toews2009sift}, offering invariance to monotonic variations in image gradient. 

Our GPU-optimized 3D SIFT-CNN method was first described and used in the context of brain MRI analysis~\citep{pepin2020large}. The original 3D SIFT-Rank method has been used in a variety of keypoint applications analyzing 3D images of the human body. These include modeling the development of the infant human brain over time~\citep{toews2012feature}, keypoint matching between different image modalities having multi-modal, non-linear intensity relationships~\citep{toews2013feature}, robust alignment of lung scans~\citep{gill2014robust}, kernel density formulation for efficient memory-based indexing from large image datasets~\citep{toews2015feature}, efficient whole-body medical image segmentation via keypoint matching and label transfer~\citep{wachinger2015keypoint,wachinger2018keypoint}, alignment of 4D cardiac ultrasound sequences~\citep{bersvendsen2016robust}, alignment of 3D ultrasound volumes for panoramic stitching~\citep{ni2008volumetric}, identifying family members from brain MRI~\citep{toews2016siblings}, and large-scale population studies using multiple neurological MRI modalities ~\citep{kumar2018multi}. The 3D SIFT-Rank method was also used for robust image alignment in the context of image-guided neurosurgery~\citep{luo2018feature}, including non-rigid registration~\citep{machado2018non} based on regularization of deformation fields via thin-plane splines and finite element models~\citep{frisken2019preliminary,frisken2019comparison}, and robust filtering of image-to-image correspondences using the variogram~\citep{luo2018using}. Alternative keypoint descriptors have been proposed based on vector data, e.g. diffusion MRI histograms of the human brain~\citep{chauvin2018diffusion}. The Jaccard distance between feature sets was introduced to characterize the bag-of-feature manifold, to automatically flag errors in large public MRI datasets~\citep{chauvin2019analyzing,chauvin2019neuroimage}, and most recently the first approach to identify family members from brain MRI~\citep{chauvin2021efficient}.

In terms of keypoint descriptors, a number of 2D descriptors have been proposed, generally based on local image gradient orientation information, including gradient orientation~\citep{Lowe}, ORB~\citep{rublee2011orb}, BRIEF~\citep{calonder2010brief}, typically these have not been extended to 3D. Rank-order normalization has been shown to improve upon standard 2D gradient descriptors~\citep{toews2009sift}. Deep learning has been used to extract keypoints and descriptors, learned invariant feature transform (LIFT)~\citep{Yi2016LIFT:Transform}, DISK~\citep{Tyszkiewicz2020DISK:Gradient}, LF-Net~\citep{Ono2018LF-Net:Images}, SuperPoint~\citep{Detone2018SuperPoint:Description}, Hardnet ~\citep{mishchuk2017working}. Surprisingly, variants of the original SIFT histogram descriptor including SIFT-Rank~\citep{Toews2009}, DSP-SIFT~\citep{Dong2015} or RootSIFT~\citep{Arandjelovic2012ThreeRetrieval} are still competitive in terms of keypoint matching performance~\citep{Balntas2017HPatches:Descriptors, Schonberger2017ComparativeFeatures}, particularly for non-planar objects and image retrieval~\citep{bellavia2020there}. SIFT keypoint extraction is based solely on fundamental, symmetric and uniform mathematical operators, e.g. the Gaussian, Laplacian, and uniformly sampled gradient operators. The resulting keypoints thus represent a class of highly informative patterns that are invariant to image scaling and rotation in addition to translation, and may be identified in any context with no explicit training procedure and nor bias towards specific training datasets used.

In this paper, we describe an efficient implementation of the 3D SIFT-Rank algorithm via GPU parallelization of the operations required to generate the Gaussian scale-space. In particular, we target the Gaussian convolution, difference of Gaussians, sub-sampling and 4D extraction. This work was the first published GPU implementation of the 3D SIFT algorithm, first validated in the context of brain image indexing~\citep{pepin2020large}. Previously, GPU processing was used to speed up the SIFT algorithm for 2D image data~\citep{heymann2007sift}, in real-time~\citep{lalonde2007real} and video processing contexts~\citep{fassold2015real}, and high-dimensional feature matching~\citep{garcia2010k} applications, and for use on mobile devices~\citep{rister2013fast}. We here provide a detailed descriptions of the implementation, justifying our parameter choices based on the CUDA API, and describing our optimisation strategy and its effect on speed. We also describe and test computationally efficient 3D descriptors including the Binary Robust Independent Elementary Features (BRIEF) approach of Calonder et al.~\citep{calonder2010brief}, and a propose a novel descriptor based on BRIEF that we call RRIEF (Ranked Robust Independent Elementary Features). These descriptors are compared to the 3D SIFT-Rank descriptor of Toews and Wells~\citep{toews2009sift,toews2013efficient} in keypoint matching experiments of magnetic resonance image (MRI)s of the human brain.

\section{Materials \& Methods}

In this section we describe our GPU optimization of the 3D scale-invariant feature transformer for keypoint extraction, the BRIEF-style descriptors extracted, and the data sets used for the experiments.

\subsection{SIFT Convolutional Neural Network}

\begin{figure}[H]
        \begin{center}
        \includegraphics[width=0.90\textwidth]{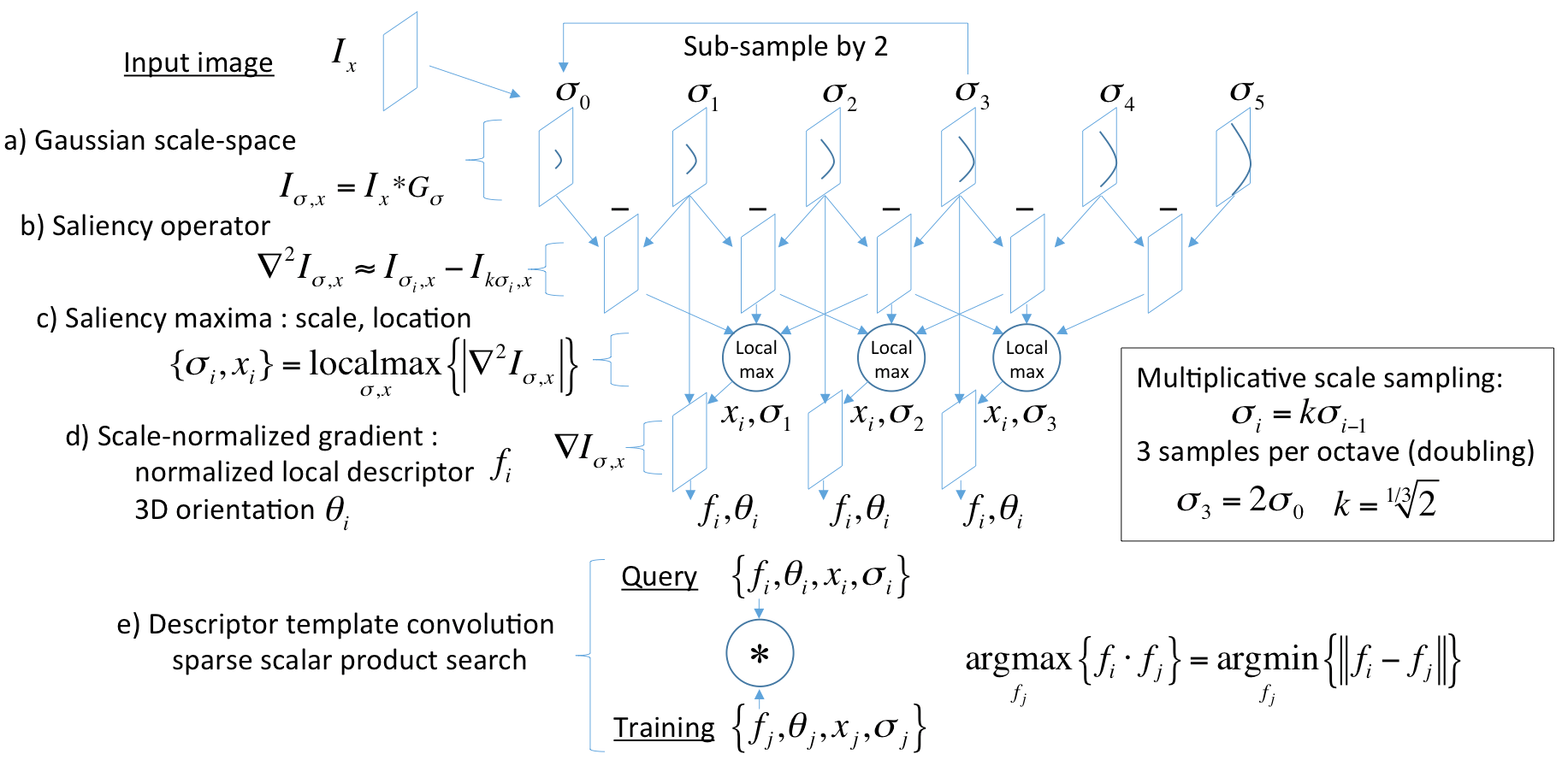}\\
        \end{center}
    \caption{Illustrating SIFT as a CNN.}
    \label{fig:suft-cnn}
\end{figure}

Here we provide a description of the scale-invariant feature transform (SIFT) algorithm, which may be viewed as a convolutional neural network (CNN), as shown in Figure~\ref{fig:suft-cnn}. The input is an image $I_x$ with scalar intensity $I \in R^1$ and sampled in 3D space $x \in R^3$. a) A single-channel scale-space $I_{\sigma,x} = I_x * G_\sigma$ is generated via Gaussian convolution, where $\sigma \in R^1$ is the stdev of the Gaussian, representing the diffusion of image information as molecules in Brownian motion or heat in the heat equation. b) The Laplacian-of-Gaussian operator serves as a self-attention mechanism reminiscent of center-surround processing in the mamalian visual system, it is approximated via a difference-of-Gaussian (DoG) operation. c) Scale-space coordinates $(\sigma_i,x_i)$ representing local maxima of attention are identified as keypoint regions. d) 3D gradient operators within keypoint regions are used to determine local 3D orientation and to form keypoint appearance descriptors. Finally, e) keypoint descriptors $f_i,f_j$ are normalized to unit length and stored in memory, and can indexing by minimizing Euclidean distance (equivalent to nearest neighbor search) or maximizing the scalar dot product (equivalent to template convolution).

\subsection{GPU Optimization}

CUDA is well known by sectors of the scientific community that commonly use GPU computation. This programming interface facilitates leveraging the Graphics Processing Unit by making the flow of memory to/from GPU possible, as well as computation using the GPU. We briefly describe the CUDA architecture and the parameters in our optimization procedure for each of the operations that we seek to optimize.

\subsubsection{CUDA architecture and parameters}

\begin{figure}[H]
  \begin{minipage}[t]{1\textwidth}
  \begin{minipage}[b]{0.465\textwidth}
    \caption{
    Computational architectures ~\citep{UPEM:GPU}~\citep{UPEM:CPU}.\textit{(a)} A CPU with a controller (yellow) and arithmetic and logic units (ALUs, green) that can be parallelized and access global memory (DRAM, red).\textit{(b)} A GPU with N parallel multiprocessors (yellow), each equipped with M parallel processors or threads (green) with individual and shared memory (red). 
    }
    \label{fig:GPU_CPU}
    \begin{center}
      \includegraphics[width=.7\textwidth]{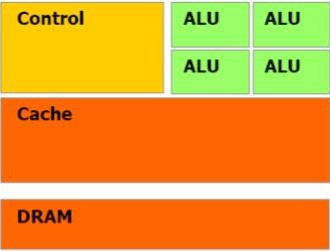}\\ 
  \end{center}
  \vspace{0.2cm}
      \begin{center}
        a) CPU
        \end{center}
  \end{minipage}
  \begin{minipage}[b]{0.475\textwidth}
    \includegraphics[width=.96\textwidth]{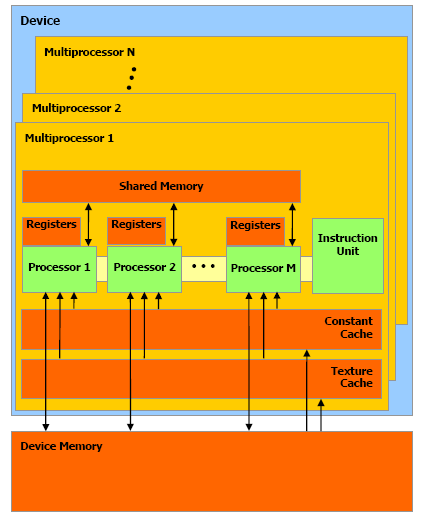}\\
    \vspace{0.2cm}
    \begin{center}
    b) GPU
    \end{center}
  \end{minipage}
  \end{minipage}
\end{figure}

In Figure~\ref{fig:GPU_CPU}, we show the CPU and GPU architectures. In a CPU, multiple arithmetic and logic units (ALUs) can work in parallel, and access a shared global memory. A GPU has several multiprocessors that can work in parallel, and each of which has multiple parallel processors that are equipped with individual and shared memory. 

CUDA is able to virtualize many blocs (multiprocessors) in each of the  $x$, $y$, and $z$ image dimensions. Each bloc is made of multiple threads (processors) coupled with a shared memory. This memory, closer to the computation units, is faster than the device memory. The CUDA API is limited to 1024 threads per bloc and to 48KB in each shared memory (Compute capability $<$ 7.0). 

Inside the CUDA kernel, the block dimension and grid dimension are two important parameters for maximizing time performance. To determine the best (x, y, z) block dimensions, we have investigated 6 models: Block dimension = $(k, k, k)$ with grid dimension $(a/k, b/k, c/k)$ for k = 1, 2, 5, 9, and 10, where $a$, $b$, and $c$ are the volume dimensions in the $x$, $y$, and $z$ directions, respectively.  We also investigated a 6th model with block dimension = (10, 10, 10) and with index modification leading to volume rotation. Model 6 is used to determine whether the CUDA environment has improved performance compared to CPU computation, particularly due to using index rotation between each convolution filter.\\
Since the convolution operation is the most time consuming one, we used it in order to select the optimal model. We thus ran each of these 6 models on the GPU convolution operation and recorded the computation time for the entire data-set for the whole SIFT operation. We then computed mean times for each operation. Finally, only the fifteen first operations were kept in order to ensure good graphical representation for each octave. We also record computation time for the maximum size volume and maximum filter size. The computation times are shown in Table \ref{tab:comp_times_table}. Based on these experiments, we select model 5 as it results in the fastest computation times.

\begin{table}[H]
    \begin{center}
        \begin{tabular}{|c|c|}
            \hline
            \makecell{Model} & Computation time\\[0.2cm]
            \hline
             1 & $ \approx 12 \times 10 ^{-2} seconds $\\[0.2cm] \hline
             2 & $\approx 2 \times 10 ^{-2} seconds $\\[0.2cm] \hline
             3 & $\approx 1.5 \times 10 ^{-2} seconds $  \\[0.2cm] \hline
             4 & $\approx 1.2 \times 10 ^{-2} seconds $ \\[0.2cm] \hline
             5 & $< 1.1 \times 10 ^{-2} seconds $\\[0.2cm] \hline
             6 & $\approx 1.2 \times 10 ^{-2} seconds $ \\[0.2cm] \hline
        \end{tabular}
        \caption{
    Computation times for each model.
    }
    \label{tab:comp_times_table}
    \end{center}
\end{table}

Below, we describe each of the operations that we are optimizing, and the parameters considered.

\subsubsection{Operation-specific optimization}

We describe the parameter setup and the kernel operations performed for each of the operations.

\begin{figure}[H]
        \begin{center}
        \includegraphics[width=0.90\textwidth]{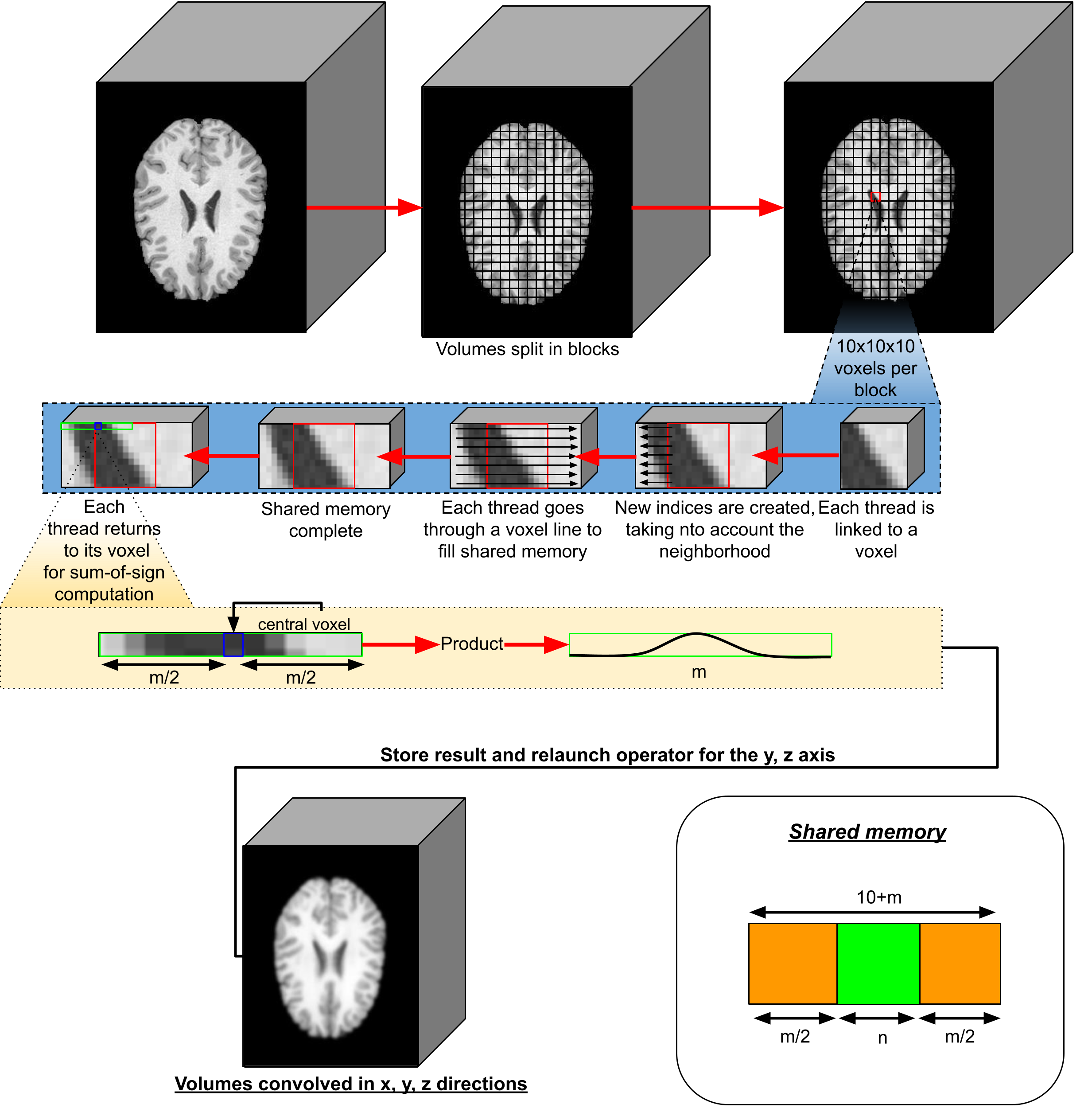}\\
        \end{center}
    \caption{Illustrating the processing for Gaussian convolution using CUDA. }
    \label{fig:Gauss_Conv}
    
\end{figure}

\paragraph{Gaussian Convolution (Figure~\ref{fig:suft-cnn} a)}
Gaussian filters operate as low pass filters, removing high-frequency intensity patterns above a cutoff threshold determined by the filter standard deviation $\sigma$.\\
The Gaussian operation is defined as : 
\begin{center}
\begin{math}
g(x, y, z) = f(x, y, z) \times \omega
 \left \{
   \begin{array}{l l l}
      g(x, y, z)  \Rightarrow the\, filtered \,volume\\
      f(x, y, z)   \Rightarrow the\, initial\,volume \\
      \omega \Rightarrow the\,filter\, applied\,to\,the\,volume,\, \omega \hookrightarrow  \mathcal{N}(\mu,\,\sigma^{2})
   \end{array}
   \right\}
\end{math}
\end{center}
Gaussian filtering is the most time consuming operation of the 3D SIFT algorithm. We implement it via three separable 1D convolutions in the $x,y,z$ spatial dimensions for efficiency. 
\begin{description}
    \item[Parameter Setup:] ~\\Before launching the kernel, we set  up the number of threads by block and number of blocks by grid based on model 5 parameters, i.e. a block dimension of (10, 10, 10) and a grid dimension of $(10/a, 10/b, 10/c)$. \\ We determine how much shared memory is required by block prior to assigning the shared memory size. We need, in addition to the  block dimension in the $x$ direction, memory equivalent to the filter size. We therefore set the dimensions of shared memory to $(10+m, 10, 10)$ with $m$ being the filter size.
    \item[Kernel operations:] ~\\These operations are illustrated in Figure \ref{fig:Gauss_Conv}. Once the kernel is launched, the image will be split into blocks. Inside each block, each thread will be linked to a voxel using its index. We check if the index  corresponds to a voxel of the volume. If not, the thread is released.\\Because the shared memory dimension is larger than the block dimension, only 10*10 threads will be used to fill it. The plane will move along the image to store each voxel and the neighbors needed for convolution into shared memory.\\ Finally, each thread is returned to the target voxel in shared memory and convolution is computed.
\end{description}

\paragraph{Sub-sampling}
Sub-sampling here refers to reducing the image size by a factor of 2 in all dimensions. According to the NVIDIA CUDA documentation, the best way of working on GPU is to divide processing in small amounts of work shared across a large number of threads, rather than vice versa.
\begin{description}
    \item[Parameter Setup :] ~\\We set the number of threads by block to (10, 10, 10), the shared memory size to (20, 20, 20) and the grid dimensions to $(a/10, b/10, c/10)$.
    \item[Kernel operations :] ~\\Two kinds of indices will be set, the first for the input volume, and the second for the output volume. We check if the indices are linked to voxels of the volumes, otherwise the thread is released. The shared memory will be filled with the input volume. Once done, 8 voxels from shared memory will be used to compute one voxel in the output volume.
\end{description}

\paragraph{Difference-of-Gaussian (Figure~\ref{fig:suft-cnn} b)}
The Difference-of-Gaussian (DoG) operation requires computing the intensity difference between two volumes. The DoG is defined as $DoG(x,y,z,\sigma)=I(x,y,z,\sigma)-I(x,y,z,\kappa\sigma)$, where $\kappa$ is a multiplicative sampling rate used in generating the scale-space, approximating the Laplacian-of-Gaussian operator $I(x,y,z,\sigma)-I(x,y,z,\kappa\sigma) \approx \nabla^2I(x,y,z,\sigma) \nabla^2 I(x,y,z,\sigma)$. Equivalent voxel coordinates between the two volumes are linked by indices, and the voxel intensity difference is set at the equivalent coordinates in the output.
\begin{description}
    \item[Parameter Setup :] ~\\No shared memory is needed for this step as we believe that filling the shared memory will lead to more cost than simple reading.
    Block dimensions are set to $(10, 10, 10)$ and the grid dimensions are set to $(a/10, b/10, c/10)$.
    \item[Kernel operations :] ~\\We begin the operation with index generation, then we check if the index is linked to a voxel in the volume, otherwise the thread is released. The output volume will then contain the intensity difference between two volumes.
\end{description}

\begin{figure}[H]
        \begin{center}
        \includegraphics[width=0.90\textwidth]{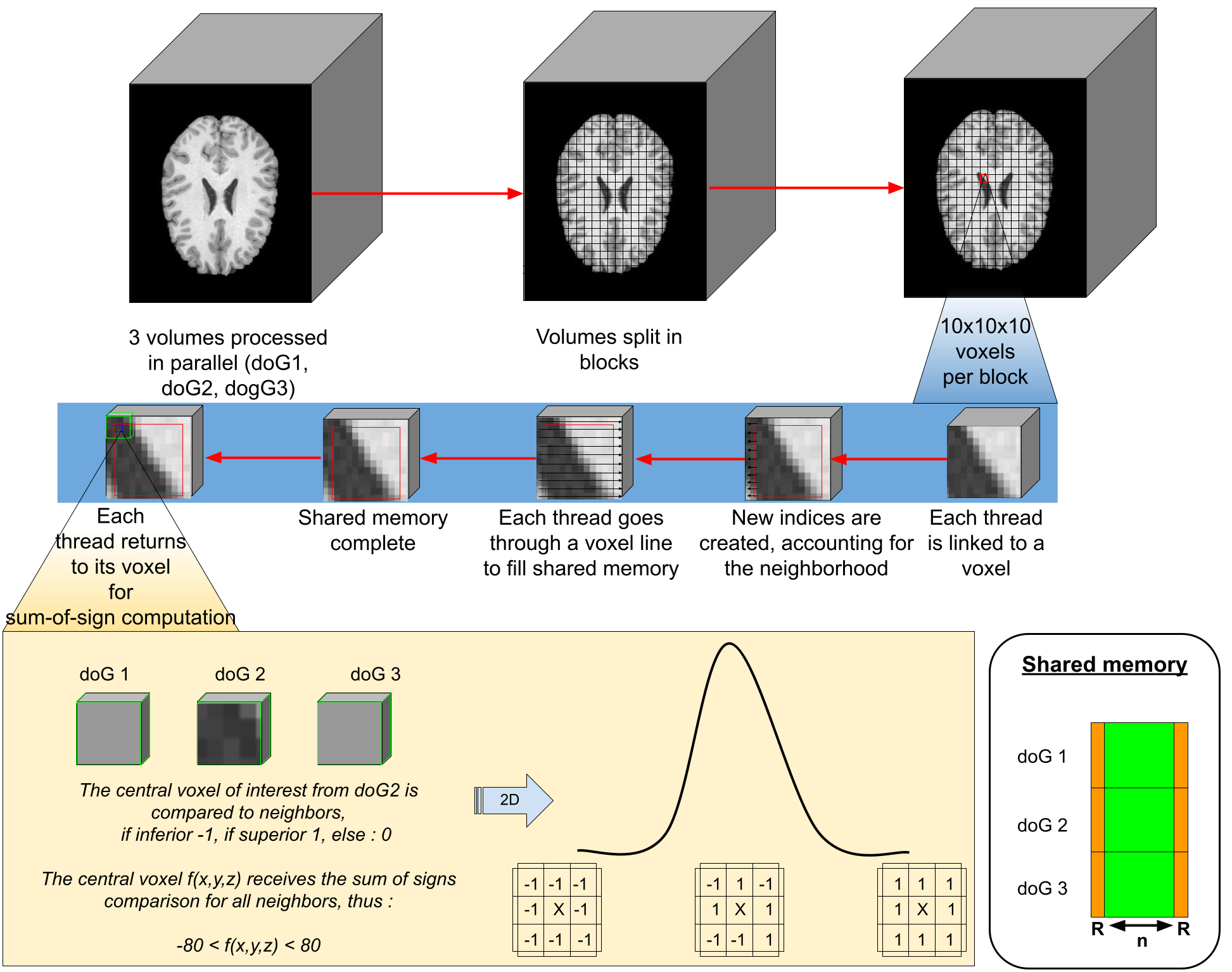}\\
        \end{center}    \caption{Illustrating the process of 4D peak extraction via the sum-of-signs operation.}
        \label{fig:Peak_extract}
\end{figure}

\paragraph{4D Peak Detection (Figure~\ref{fig:suft-cnn} c)}
4D extrema detection involves identifying either peaks or valleys of the DoG saliency operator $local~\underset{x,y,z,\sigma}{\arg\max}\{ \| DoG(x,y,z,\sigma) \| \}$, approximating maxima of the LoG operator $local~\underset{x,y,z,\sigma}{\arg\max}\{ \| \nabla^2I(x,y,z,\sigma) \| \}$ which represent informative keypoint coordinates in scale-space. This is performed using three DoG volumes computed at consecutive scale $\sigma$ increments, identifying voxel locations that are either higher or lower than all $27+26+27=80$ adjacent voxels. The method used here is inspired by the environment used in computing the 2D LBP descriptor~\citep{LBP94,LBP96}.
\begin{description}
    \item[Parameter Setup :]~\\Block dimensions are again set to $(10, 10, 10)$ and grid dimension are set to \newline$(a/10, b/10, c/10)$. Three DoG volumes are required for this operation; we thus need enough space for 3 blocks of volumes and their neighbors in shared memory. Shared memory is set to 3$\times$(12, 12, 12).
    \item[Kernel operations :] ~\\These operations are illustrated in Figure \ref{fig:Peak_extract}. Indices are created to fill the shared memory of the 3 voxel blocks. Once this operation is finished, the sum of sign function is applied to each voxel. This function is defined by: 
    \begin{center}
\begin{math}
f(x\textsubscript{i}, \sigma \textsubscript{i}) = \sum_{i, j\in N} sign(I(x\textsubscript{i}, \sigma \textsubscript{i})-I(x\textsubscript{j}, \sigma \textsubscript{j})) 
 \left \{
   \begin{array}{l l l}
      x\textsubscript{i}  \Rightarrow target\,voxel \\
      \sigma\textsubscript{i}  \Rightarrow DOG\,i \\
      x\textsubscript{j}  \Rightarrow neighbors\,voxels \\
      \sigma\textsubscript{j}  \Rightarrow DOG\,i-1,\,i,\,i+1 \\
      I\textsubscript{i}  \Rightarrow volume \\
   \end{array}
   \right\}
\end{math}
\end{center}
The results are stored in an output extrema map.
    \item[Post-operations :] ~\\After creating the extrema map, peaks and valleys are identified as values of 80 and -80, respectively. The possibility of modifying this interval has been set up to allow extraction of keypoints that are not strictly local extrema.
\end{description}
\subsubsection{FAST 3D Descriptors (Figure~\ref{fig:suft-cnn} d)}

Here we investigate several alternative descriptors based on the well-known 2D BRIEF descriptor~\citep{calonder2010brief}, where each descriptor element is defined by a binary comparison between intensities $I(p_1)$ and $I(p_2)$ at two different point locations $p_1$ and $p_1$. We investigate the result of descriptor normalization and point pair selection strategies.

The two normalization strategies are as follows:
\begin{description}
    \item[BRIEF :]~\\ A patch of volume is selected and blurred to increase description of intensity information. Then $n$ pairs of points are subtracted and the result is binarized.
    \item[RRIEF :]~\\In an attempt to increase the effectiveness of BRIEF, we propose the RRIEF descriptor, similar to BRIEF except that the result of each difference is a floating point number and the entire descriptor set is rank-ordered as in case of the SIFT-Rank descriptor.
\end{description}

\begin{figure}[H]
        \begin{center}
        \includegraphics[width=0.90\textwidth]{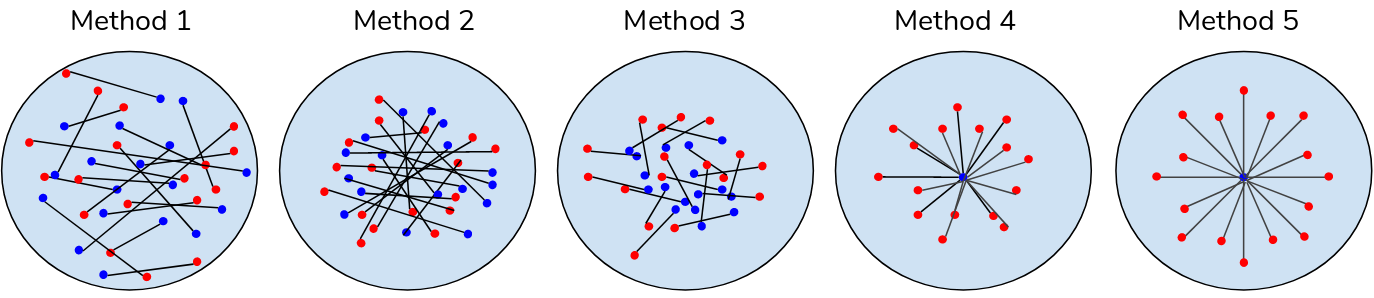}\\
        \end{center}
    \caption{Illustrating the five point selection methods for BRIEF-inspired descriptors, methods 1, 2, 3 and 5 were proposed by Calonder et al.~\citep{calonder2010brief}, method 4 is novel in this study.}
    \label{fig:point-selection}
\end{figure}

For each normalization strategy, five point selection strategies are considered to identify a set of $n$ pairs of 3D points $(p_1,p_2)$, $p_i=\{x_i,y_i,z_i\}$ for binary comparisons. Four of these were described in the 2D descriptor context~\citep{calonder2010brief}, and a novel fifth method is investigated here (shown as method 4 below). These are illustrated in 2D $x,y$ coordinates in Figure~\ref{fig:point-selection}: 
\begin{description}
\item[Method 1 :] Points $p_1,p_2$ are sampled from a uniform distribution with a circle of radius $2\sigma$.
\item[Method 2 :] Points $p_1,p_2$ are sampled from normal density $N(0,\sigma)$ with mean zero and standard deviation $\sigma$.
\item[Method 3 :] Points $p_1$ is sampled from normal density $N(0,\sigma)$ and $p_2$ from normal density $N(p_1,\sigma)$ centered upon $p_1$.
\item[Method 4 :] Points $p_1$ are located in the center of the patch and points $p_2$ are sampled from normal density $N(0,\sigma)$.
\item[Method 5 :]  Points $p_1$ is located in the center of the patch and points $p_2$ are distributed over a regular polar coordinate grid.
\end{description}

\begin{figure}[H]
        \begin{center}
        \includegraphics[width=0.70\textwidth]{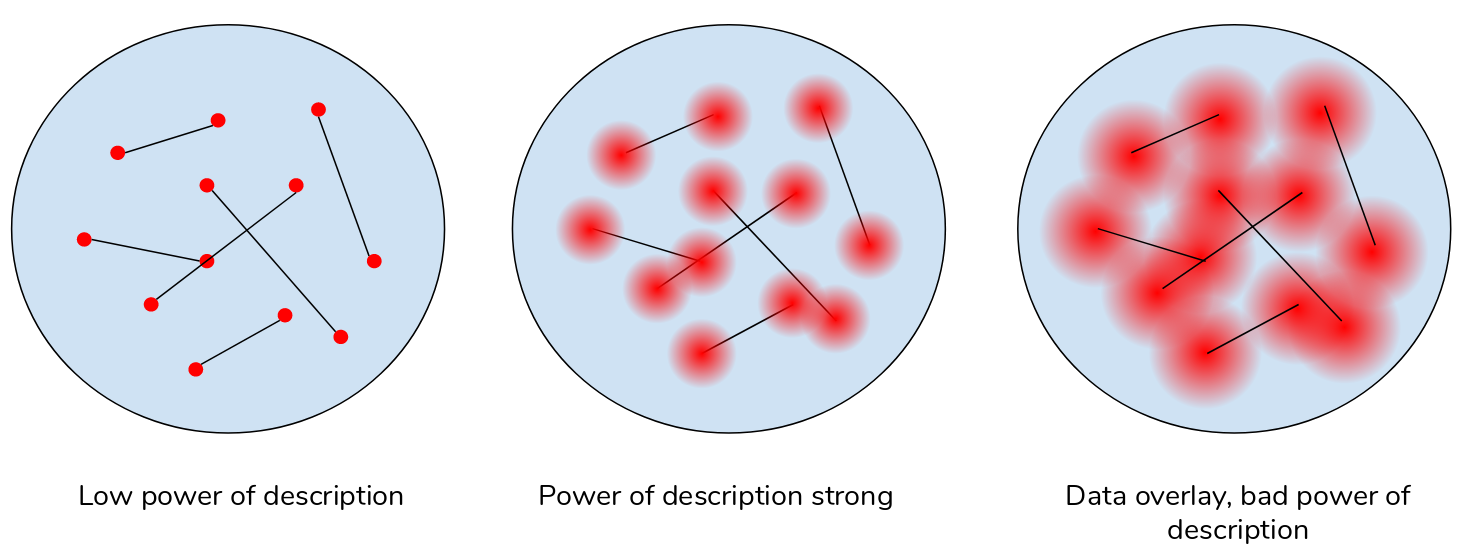}
        \end{center}
    \caption{Illustrating the effect of pre-blurring, where an intermediate value of Gaussian blurring (center) is optimized for effective description of point pair differences.}
    \label{fig:power}
\end{figure}

Prior to point pair difference computation, a Gaussian blur is applied to the patch in order to reduce noise and improve the descriptor performance. The Gaussian sigma blur parameter must be chosen for optimal performance as shown in Figure~\ref{fig:power}, i.e. large enough to reduce intensity variations due to noise but small enough to maintain informative image signal.

\subsection{Experiments and Data}

Experiments seek to investigate and quantify the computational efficiency and the descriptor matching accuracy, based on 3D brain volumes. Computational efficiency is assessed in terms of mean computation time for 3D SIFT feature extraction from individual images, for generating three types of descriptors: the original SIFT-Rank~\citep{toews2013efficient} and novel BRIEF and RRIEF descriptors proposed here. Descriptor accuracy is assessed via the sum of matching features resulting from nearest neighbor descriptor matching trials between keypoints extracted in pairs of brain volumes. A variant of the Hough transform to identify a robust 7 degree-of-freedom transform (i.e. 3D rotation, translation and isotropic scale) aligning pairs of 3D volumes~\citep{toews2013efficient}, matching keypoint pairs are identified as inliers to the global Hough transform. Matching results are provided for proposed BRIEF and RRIEF descriptor parameters including 5 normalization strategies and 10 different sigma values.

Experiments were performed on volumes from the OASIS1 data set~\citep{marcus2007open}, an open access database of T1-weighted MRI volumes of the human brain, including healthy subjects and individuals affected by Alzheimer's disease and natural aging. Table~\ref{tab:dataset} contains demographic and statistical information for this dataset.
\begin{table}[H]
    \begin{center}
        \begin{tabular}{|c|c|c|c|c|c|c|}
            \hline
            \makecell{Data-set} & Subjects & \makecell{Gender \\ (M/F)} & \makecell{Age\\(Min/Avg/Max)} & \makecell{Available\\Volumes} &\makecell{Voxel size\\(mm)} & \makecell{Key-points\\(Avg/Volume)}\\[0.2cm]
\hline
            OASIS 1 & 416 & 160 / 256 & 18 / 53 / 96 & 5 & 1.0 & 2896\\[0.2cm]
            \hline
        \end{tabular}
        \caption{
    Data-set demographic and statistical information for the OASIS
    }
    \label{tab:dataset}
    \end{center}
\end{table}

\section{Results}

\subsection{Gaussian Convolution}
\begin{figure}[H]
        \begin{center}
        \includegraphics[width=0.80\textwidth]{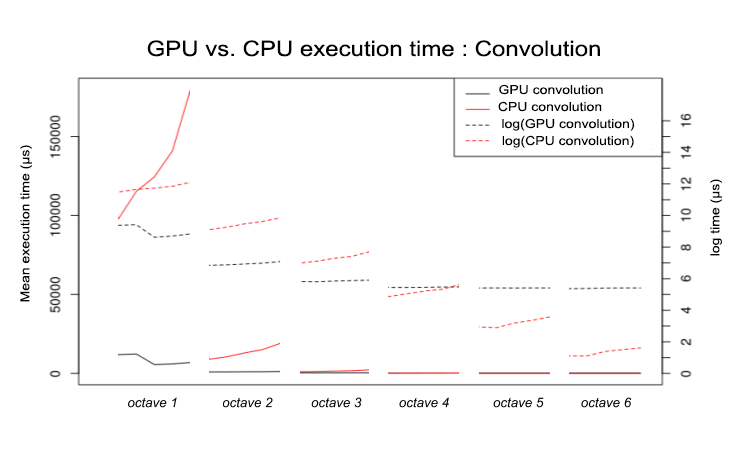}\\
        \end{center}
    \caption{Speedup for the Gaussian convolution operation, comparing GPU vs. CPU for 6 octaves. Note that there are 6 convolutions per octave with increasing filter size.}
    \label{fig:gauss_cpu_vs_gpu}
\end{figure}

Computational complexity in the SIFT algorithm is dominated by the Gaussian convolution operation. In Figure~\ref{fig:gauss_cpu_vs_gpu} we can observe maximum execution times of up to \begin{math}
18 \times 10 ^{-2} 
\end{math} seconds for CPU convolution, and \begin{math}
1.1 \times 10 ^{-2} 
\end{math} seconds for GPU convolution. Note that the CPU implementation is affected by the filter size, which is largest in the final convolution of each octave. \\Each octave (volume sub-sampling by a factor of 2) reduces the CPU time by less than 8.5 and the GPU time by less than 6.2.\vspace{0.5cm}\\
In logarithmic units, we can see for CPU : 
\begin{itemize}
    \item maximum time of 12 log(mean of time in $\mu$seconds)
    \item minimum time of less than 2 log(mean of time in $\mu$seconds)
\end{itemize}
and for GPU : 
\begin{itemize}
    \item maximum time of 10 log(mean of time in $\mu$seconds)
    \item minimum time of less than 5 log(mean of time in $\mu$seconds)
\end{itemize}
\subsection{Sub-sampling}
\begin{figure}[H]
        \begin{center}
        \includegraphics[width=0.80\textwidth]{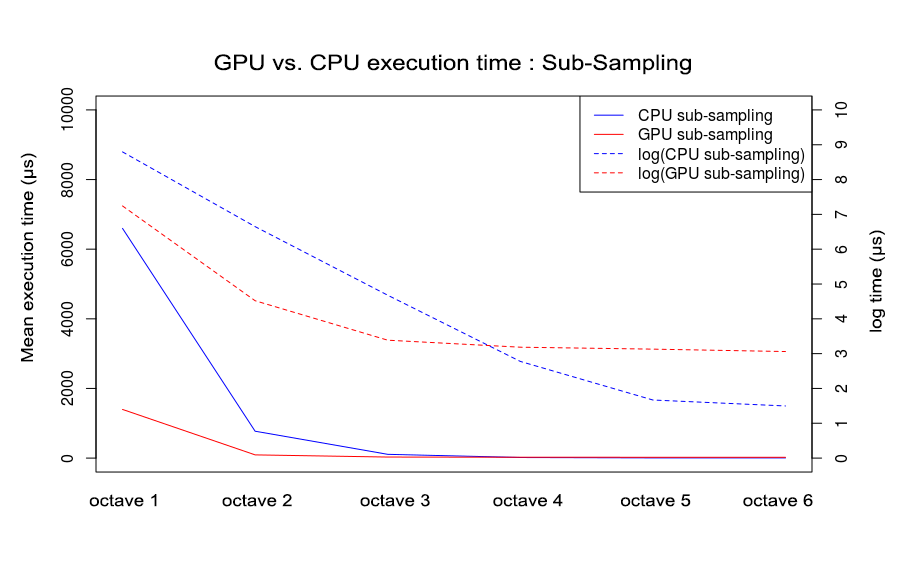}\\
        \end{center}
    \caption{Speedup for the sub-sampling operation, comparing GPU vs. CPU for 6 octaves.}
    \label{fig:Sub_sample}
\end{figure}

The sub-sampling time comparison is shown in Figure~\ref{fig:Sub_sample} for six octaves, note sub-sampling is performed once per octave. The maximum execution times are approximately \begin{math}
6.6 \times 10 ^{-3} 
\end{math} seconds for the CPU vs. \begin{math}
1.3 \times 10 ^{-3} 
\end{math} seconds for the GPU implementation. Between the first and second sub-sampling on CPU the time is reduced by a factor of 8.55 vs. 15.21 for GPU. In logarithmic units, we can see for CPU : 
\begin{itemize}
    \item maximum time of 8.7 log(mean of time in $\mu$seconds)
    \item minimum time of less than 1.4 log(mean of time in $\mu$seconds)
\end{itemize}
and for GPU : 
\begin{itemize}
    \item maximum time of 7.2 log(mean of time in $\mu$seconds)
    \item minimum time of more than 3 log(mean of time in $\mu$seconds)
\end{itemize}
\subsection{Difference-Of-Gaussian}
\begin{figure}[H]
        \begin{center}
        \includegraphics[width=0.80\textwidth]{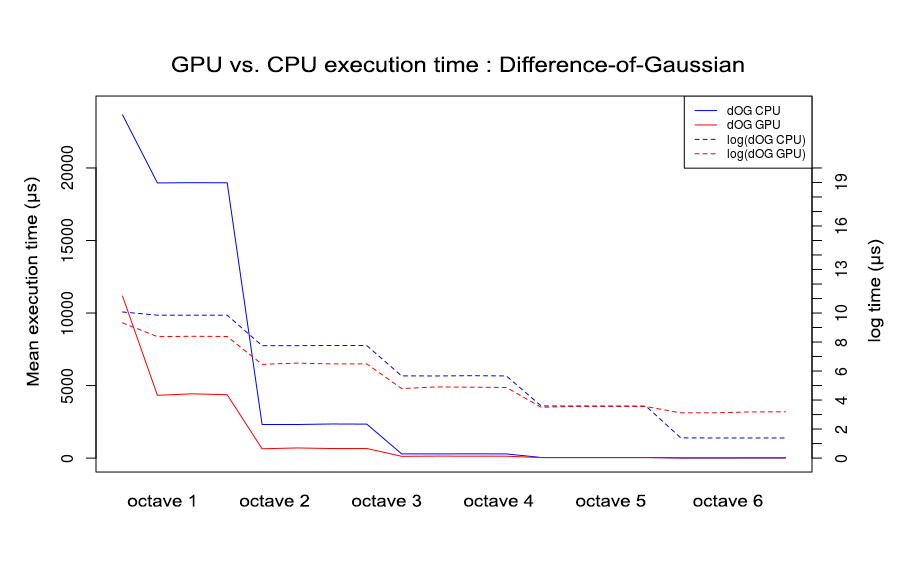}\\
        \end{center}
    \caption{Speedup for the difference-of-Gaussian operation, comparing GPU vs. CPU for 6 octaves.}
    \label{fig:DOG}
\end{figure}

For the difference-of-Gaussian (DoG), we can observe maximum execution times in Figure~\ref{fig:DOG} of around \begin{math}
2.3 \times 10 ^{-2} 
\end{math} seconds on the CPU and \begin{math}
1.1 \times 10 ^{-2} 
\end{math} seconds on the GPU. Between the 2 first DoGs, execution time is reduced by a factor of 1.2 on the CPU and 2.6 on the GPU. \vspace{0.5cm}\\
In logarithmic units, we can see for CPU : 
\begin{itemize}
    \item maximum time of 10.1 log(mean of time in $\mu$seconds)
    \item minimum time of less than 1.4 log(mean of time in $\mu$seconds)
\end{itemize}
and for GPU : 
\begin{itemize}
    \item maximum time of 9.3 log(mean of time in $\mu$seconds)
    \item minimum time of more than 3.1 log(mean of time in $\mu$seconds)
\end{itemize}
\subsection{4D Peak Detection}
\begin{figure}[H]
        \begin{center}
        \includegraphics[width=0.80\textwidth]{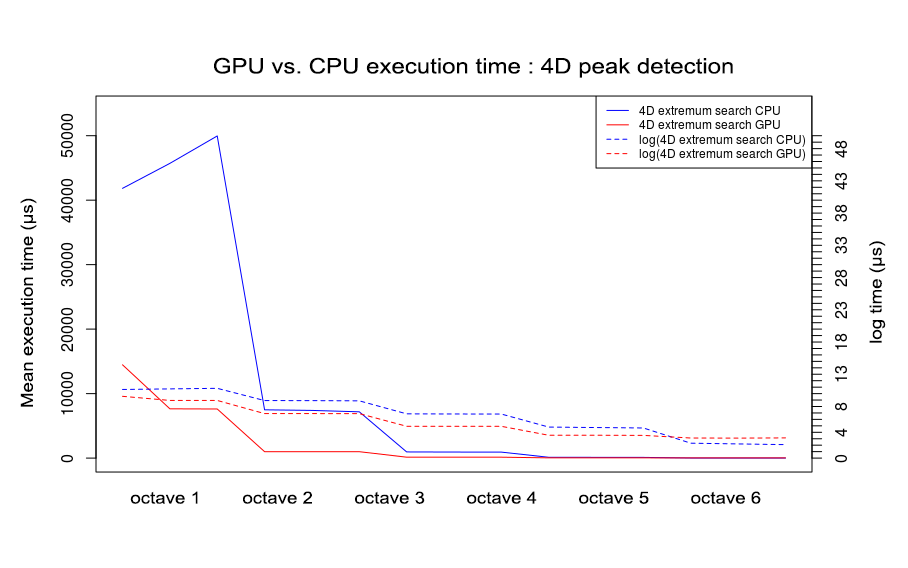}\\
        \end{center}
    \caption{Speedup achieved for 4D Peak Detection computation: GPU vs. CPU implementations.}
    \label{fig:Peak_Extract_time}
\end{figure}

Finally regarding CUDA optimisation, for the 4D peak detection operation, we can observe maximum execution times in Figure~\ref{fig:Peak_Extract_time} of approximately \begin{math}
4.9 \times 10 ^{-2} 
\end{math} seconds on the CPU and \begin{math}
1.4 \times 10 ^{-2} 
\end{math} seconds on the GPU. \\
Note that differences in computation time between the CPU and GPU implementations are partially due to algorithmic differences, where the CPU implementation is affected by the number of keypoints extracted at the highest resolution in octave 1.
\vspace{0.5cm}\\
In logarithmic units, we can see for CPU : 
\begin{itemize}
    \item maximum time of 10.6 log(mean of time in $\mu$seconds)
    \item minimum time of less than 2.1 log(mean of time in $\mu$seconds)
\end{itemize}
and for GPU : 
\begin{itemize}
    \item maximum time of 9.6 log(mean of time in $\mu$seconds)
    \item minimum time of more than 3.1 log(mean of time in $\mu$seconds)
\end{itemize}
\subsection{FAST 3D Descriptors}
\subsubsection{BRIEF and RRIEF temporal optimisation} 
\begin{figure}[H]
        \begin{center}
        \includegraphics[width=.5\textwidth]{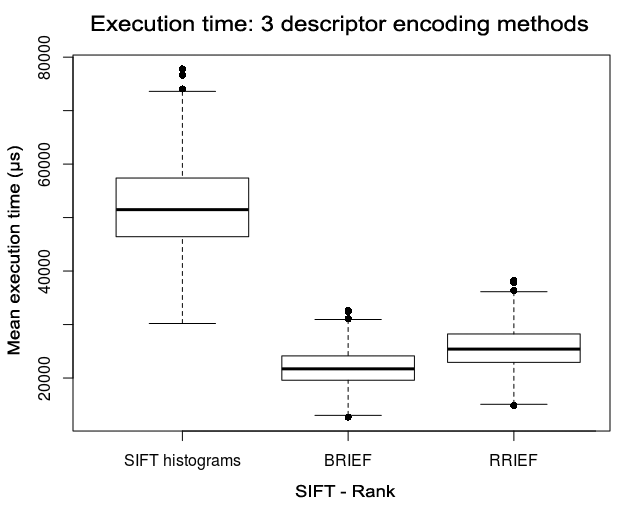}
        \end{center}
    \caption{Mean execution time for three keypoint descriptors: 3D SIFT-Rank, BRIEF, RRIEF}
    \label{fig:Desc_method_time}
\end{figure}

The plot in Figure~\ref{fig:Desc_method_time} shows the mean execution times for descriptors, including the standard 3D SIFT-Rank descriptor, and proposed fast binary BRIEF and ranked RRIEF descriptors.
For the original 3D SIFT-Rank descriptor, \begin{math}
5.1 \times 10 ^{-2} 
\end{math} seconds (min \begin{math}
3.0 \times 10 ^{-2} 
\end{math} seconds, max \begin{math}
7.8 \times 10 ^{-2} 
\end{math} seconds).\\For the 3D BRIEF descriptor adapted from Calonder et al., the mean is \begin{math}
2.2 \times 10 ^{-2} 
\end{math} seconds (min \begin{math}
1.3 \times 10 ^{-2} 
\end{math} seconds, max \begin{math}
3.3 \times 10 ^{-2} 
\end{math} seconds). \\For the ranked RRIEF descriptor, the mean is \begin{math}
2.5 \times 10 ^{-2} 
\end{math} seconds (min \begin{math}
1.5 \times 10 ^{-2} 
\end{math} seconds, max \begin{math}
3.8 \times 10 ^{-2} 
\end{math} seconds.
\subsubsection{BRIEF accuracy}
\begin{figure}[H]
        \begin{center}
        \includegraphics[width=1\textwidth]{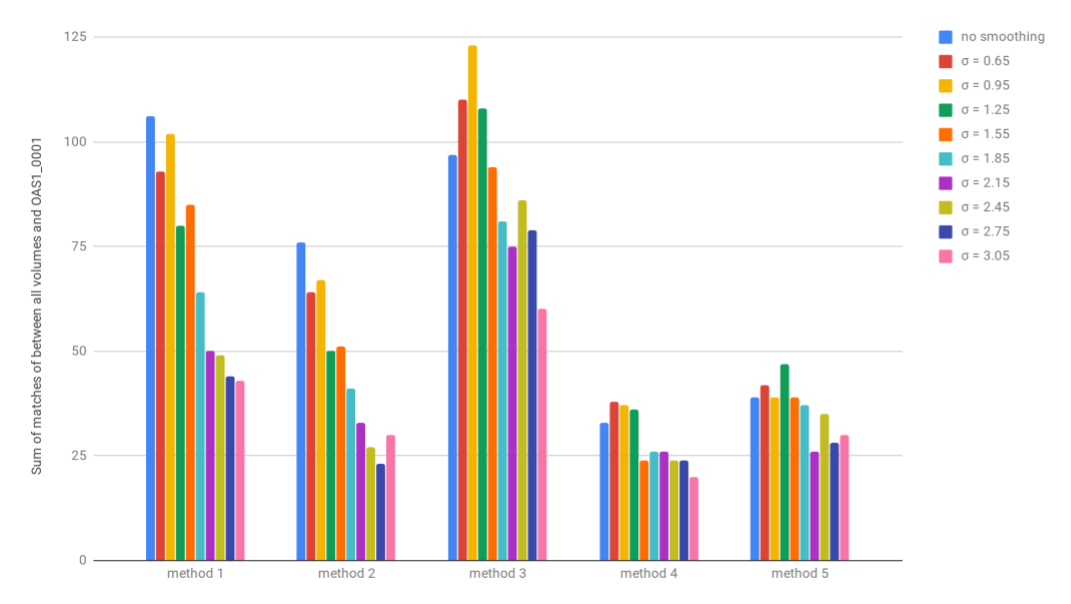}\\
        \end{center}
    \caption{Comparison of successful matches for different methods and different sigmas for the 3D BRIEF descriptor.}
    \label{fig:Desc_BRIEF_power}
\end{figure}

The accuracy of our adaptation of the 3D BRIEF descriptor was evaluated for five methods and 10 sigma values in order to identify the optimal configuration, the results are shown in Figure~\ref{fig:Desc_BRIEF_power}.
\begin{description}
\item[Method 1 :] The highest number of matches is obtained without any smoothing with 106 matches followed by 102 matches for a sigma 0.95. Sigma upper than 1.85 are bellow 50 matches.
\item[Method 2 :] No smoothing is also the best parameter with 76 matches, followed again by sigma 0.95 with 67 matches. Sigma upper than 1.55 are bellow 50 matches.
{\bf \item[Method 3 :] A sigma parameter of 0.95 gives the highest number of matches with 123, followed by 0.65 with 110 matches. Sigma 3.05 is the only one below 75 matches.}
\item[Method 4 :] A sigma parameter of 0.65 gives the highest number of matches with 38. All the sigmas are below 50 matches.
\item[Method 5 :] Sigma 1.47 gives 47 matches, this is the maximum for this method.
\end{description}

\subsubsection{RRIEF accuracy}
\begin{figure}[H]
        \begin{center}
        \includegraphics[width=1\textwidth]{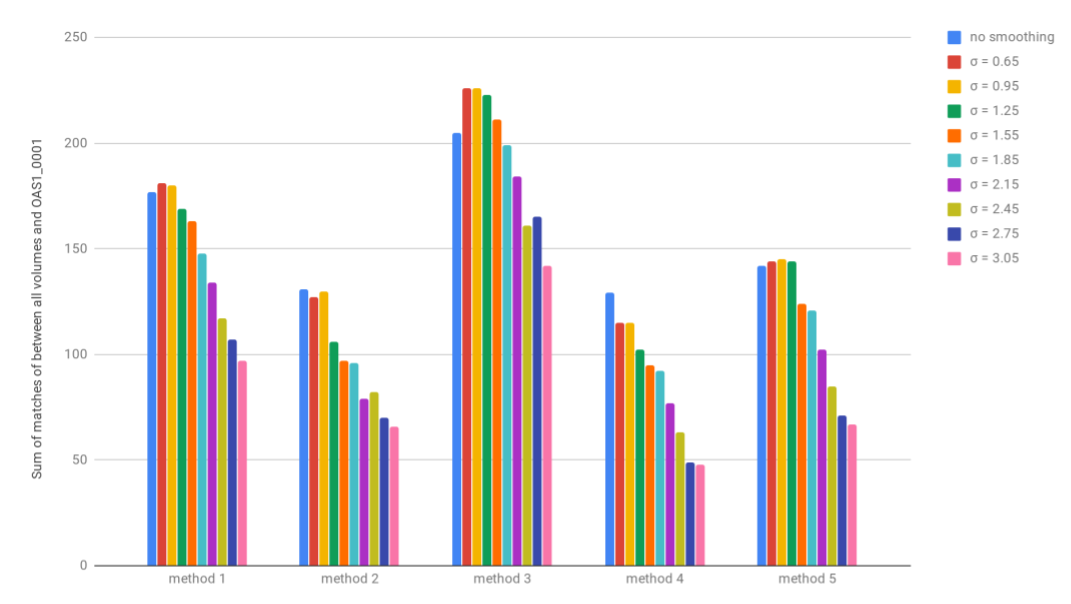}\\
        \end{center}
    \caption{Comparison of successful matches for different methods and different sigmas for the 3D RRIEF descriptor.}
    \label{fig:Desc_RRIEF_power}
\end{figure}

As for the BRIEF, five methods and 10 sigmas have been experimented to search best parameters for this new method (Figure~\ref{fig:Desc_RRIEF_power}).
\begin{description}
\item[Method 1 :] The highest number of matches is obtained for sigma 0.65 with 181 matches followed by 180 matches for sigma 0.95. Sigmas greater than 1.85 are below 150 matches. There is no sigma with less than 50 matches.
\item[Method 2 :] No smoothing is the best parameter with 131 matches, followed by sigma 0.95 with 130 matches. Sigmas greater than 1.25 are below 100 matches.
{\bf \item[Method 3 :] Sigma parameters 0.95 and 0.65 give the highest number of matches with 226 matches. Sigma 3.05 is the only one below 150 matches.}
\item[Method 4 :] No smoothing gives a better result than smoothing with any sigma. The highest number of matches is 129. Sigmas upper than 2.45 have less than 50 matches.
\item[Method 5 :] Sigma 0.95 results in 145 matches, this is the maximum for this method. Sigmas greater than 2.15 result in less than 100 matches.
\end{description}
\section{Discussion}
\subsection{Block parameters}
A decrease in volume size leads to a decrease of time. From these results we can assert that model 1 is the least efficient, and that using multiple threads leads to important time savings. The model 5 is the most efficient, at maximum 20 times faster than model 1. Model 6 is not the most efficient, indicating that GPU computation doesn't use the CPU cache storage model.

\subsection{Gaussian Convolution}
Optimizing the Gaussian convolution operation leads to the largest speedup, the GPU convolution is at maximum 20 times more efficient than CPU convolution.
CPU convolution is more efficient than GPU convolution for small volume sizes, i.e. below a resolution of (20,20,20) voxels. This is due to additional overhead required for GPU computation, including GPU to CPU copy-back action or even the CUDA environment initialization.
\subsection{Sub-sampling}
The sub-sampling operation is already relatively fast on the CPU, however GPU sub-sampling offers a speedup of approximately 3 times. As previously noted, CPU operations are more efficient for small volume sizes.

\subsection{Difference-Of-Gaussian}
The difference-of-Gaussian operation also leads to an important speedup, where the GPU operation 2 times faster than the CPU for the largest volumes. Again, the GPU operation is less efficient than the CPU for volumes below a certain minimal size. 
\subsection{4D Peak Detection}
The GPU method for peak detection offers a more important speedup than for subsampling and DoG computation, with a maximum speedup of 3 times on the GPU. This speedup here is in part due to algorithmic differences between peak detection on the CPU vs. GPU, which are brute force search on the CPU vs. accumulating a 3D extremum map using the sum-of-signs method on the GPU. Again the GPU implementation is less efficient than the CPU for small volume sizes below a threshold, e.g. (20,20,20) voxels here. A size check can be used to determine the most efficient method. This has a minor effect on overall computational time, however, which is dominated by large volumes however, and

\subsection{FAST 3D Descriptor}
The fast 3D descriptor offers a speed improvement, 3D BRIEF is 2X faster that the original SIFT-Rank, RRIEF is slightly slower due to the sorting procedure. \\
In terms of accuracy, the best sampling method was method three with optimal blur parameter sigma 0.95, for both 3D BRIEF and RRIEF. Thus among a variety of options for sampling pairs of image points $p_1,p_2 \in R^3$ to compute a binary difference $I(p_1)-I(p_2)$, the method leading to the most informative binary descriptors was to first 1) randomly sample $p_1 \sim N(0,\sigma)$ from a keypoint-centered Gaussian $N(0,\sigma)$ then 2) $p_2 \sim N(p_1,\sigma)$ conditioned by the mean of $p_1$.
The RRIEF descriptor encoding rank leads to 226 matches, twice as many matches as the 123 for the binary BRIEF descriptor, demonstrating the performance improvement afforded by ranking, at the memory cost of maintaining a vector of rank indices as opposed to a bit vector. Note that the original SIFT-Rank descriptors lead to 342 matches. Although the BRIEF method is less accurate, it requires only 1 bit per element as opposed to 6 bits for 64-element SIFT-Rank descriptor elements. Users thus have the option to trade off descriptor power vs. memory.
\section{Conclusion}
\begin{figure}[H]
        
        \begin{center}
        \includegraphics[width=.74\textwidth]{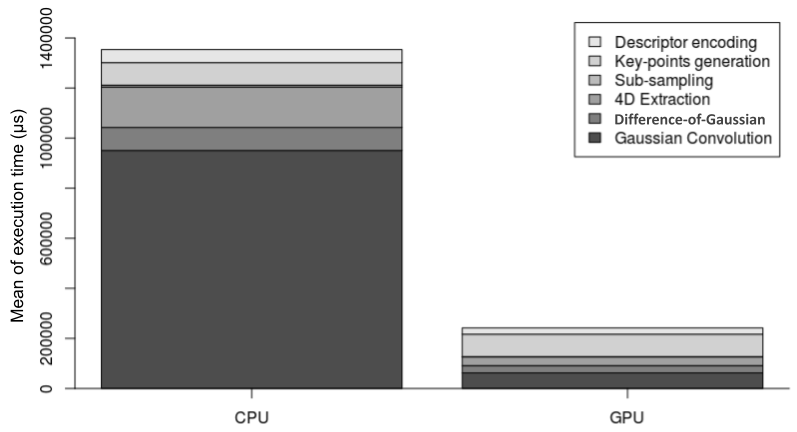}\\
        \end{center}
    \caption{Performance comparison for CPU and GPU implementations.}
    \label{fig:Conclusion_times}
\end{figure}

This study focused on the temporal optimisation of five steps of the 3D SIFT keypoint extraction pipeline, by implementing the SIFT algorithm as a convolutional neural network on the GPU using CUDA. The first four are steps in the computational pipeline for keypoint extraction and the fifth was an evaluation of a novel fast descriptor. Figure~\ref{fig:Conclusion_times} illustrates the performance improvements in comparison to the overall computational process.
These are as follows
\begin{description}
\item[Gaussian convolution]: 15$\times$ improvement for the GPU vs. CPU implementation.
\item[4D Peak detection]: 3$\times$ improvement of the GPU sum-of-signs algorithm vs the CPU method.
\item[Sub-sampling]: 4$\times$ improvement for the GPU vs. CPU implementation.
\item[DoG]: 3$\times$ improvement for the GPU vs. CPU implementation.
\item[Fast descriptor]: 2$\times$ computational speedup and 6$\times$ memory savings relative to the original 64-element SIFT-Rank descriptor. They are not as descriminative for matching, leading to 3$\times$ fewer matches in brain image matching experiments, however require 6$\times$ less memory, and thus offer a means of trading off matching power for memory footprint.
\end{description}
For several final notes, the RRIEF descriptor offers better matching performance vs BRIEF for a negligible computational increase, however RRIEF requires additional memory to store rank indices (e.g. 6 bits per descriptor element for 64-bit descriptors) as opposed to 2 bit binary BRIEF descriptors. In future work, these methods could be adapted to 2D image keypoints. Our GPU implementation of the SIFT algorithm may be viewed as a specialized convolutional neural network with filter parameters specified by the Gaussian scale-space, future work will involve parameter optimization and porting the method to standard CNN development frameworks. The 3D GPU SIFT project code may be found at the following link: \url{https://github.com/CarluerJB/3D_SIFT_CUDA}.
\bibliography{biblio}

\section*{Software used}
\begin{description}
\item[3D SIFT :]\href{http://www.matthewtoews.com/fba/featExtract1.6.tar.gz}{http://www.matthewtoews.com/fba/featExtract1.6.tar.gz}
\end{description}
\end{document}